\documentclass{article}

\usepackage{arxiv}

\usepackage[utf8]{inputenc} 
\usepackage[T1]{fontenc}    
\usepackage{hyperref}       
\usepackage{url}            
\usepackage{booktabs}       
\usepackage{amsfonts}       
\usepackage{nicefrac}       
\usepackage{microtype}      
\usepackage{graphicx}
\usepackage{multirow}
\usepackage{authblk}
\usepackage{caption}
\usepackage{subcaption}

\usepackage{mathtools}

\usepackage{xcolor}

\title{From Classification to Localization and Clinical Validation: Large-Scale Development of a Deep Learning System for Thoracic Disease Detection on Chest Radiographs in Thailand}

\author[1]{Isarun Chamveha}
\author[1]{Tretap Promwiset}
\author[1]{Napat Wanchaitanawong}
\author[3]{Trongtum Tongdee}
\author[3]{Pairash Saiviroonporn}
\author[2]{Warasinee Chaisangmongkon}
\affil[1]{Perceptra Co., Ltd., Bangkok, Thailand}
\affil[ ]{\texttt{\{isarun, tretap, napat\}@perceptra.tech}}
\affil[2]{Institute of Field Robotics, King Mongkut's University of Technology Thonburi, Bangkok, Thailand}
\affil[ ]{\texttt{warasinee.cha@mail.kmutt.ac.th}}
\affil[3]{Radiology Department, Faculty of Medicine Siriraj Hospital, Mahidol University, Bangkok, Thailand}
\affil[ ]{\texttt{\{trongtum, pairash.sai\}@gmail.com}}

\begin{document}
\maketitle

\begin{abstract}
Chest radiography (CXR) remains the most widely used thoracic imaging
modality, yet expert interpretation is constrained by a severe shortage
of radiologists in Thailand and across Southeast Asia. Local adaptation
of deep learning models to Thai data has been shown to substantially
improve accuracy on Thai populations. Here we present the development
and comprehensive validation of the chest radiograph analysis model in
Inspectra CXR version 5, a deep learning system that performs
multi-label thoracic disease classification and weakly supervised lesion
localization within a single model. The architecture couples a
DenseNet-121 backbone with Attend-and-Compare Modules (ACM) and a
Probabilistic Class Activation Map (PCAM) aggregation layer, producing a
per-condition classification score and heatmap simultaneously. The model
was developed on 874,858 frontal chest radiographs with paired
radiologist reports from Siriraj Hospital, Bangkok. On a held-out,
radiologist-verified in-domain test set of 19,871 cases, it achieved a
mean AUROC of 0.994 (mean sensitivity 92.4\%, specificity 98.6\%) across
nine clinically important conditions. On an independent generalization
set of 5,992 cases from 13 hospitals across Thailand, the mean AUROC was
0.970, indicating robust transfer across sites. For localization,
evaluated on 4,549 radiologist-annotated cases, the model attained a
mean lesion-localization fraction (LLF) of 77.9\% at 0.59 non-lesion
localizations per image. In a usability evaluation with five thoracic
radiologists, the system reached a classification concordance of 93.6\%,
a localization concordance of 94.7\%, and a mean System Usability Scale
(SUS) score of 89. These results indicate that a locally developed,
localization-capable CXR system can deliver high accuracy, generalize
across heterogeneous Thai hospitals, and earn the trust of practicing
radiologists.
\end{abstract}


\section{Introduction}

In modern clinical practice, chest X-ray (CXR) is one of the most widely used methods for diagnosing abnormal conditions in the chest and nearby structures, owing to its noninvasive nature, low cost, and high availability. Chest radiographs can reveal abnormalities in the lung parenchyma, mediastinum, rib cage, and heart, allowing physicians to determine the causes of various illnesses and monitor treatment for a range of life-threatening diseases such as pneumonia, tuberculosis, and cancer. In Western societies, CXR is performed on average 236 times per 1,000 patients, accounting for $25\%$ of all diagnostic imaging procedures \cite{speets2006chest}.

In Thailand and neighboring Southeast Asian countries, thoracic diseases constitute the leading causes of death across all age groups. Among the ten leading causes of death in Southeast Asia \cite{rao2013mortality}, five are thoracic abnormalities --- tuberculosis, lung cancer, pneumonia, heart disease, and chronic obstructive pulmonary disease --- all of which rely on chest X-ray imaging as a primary method of diagnosis. As the region has the highest rate of tuberculosis (TB) according to the World Health Organization \cite{whotbreport2019} and accounts for $44\%$ of TB cases worldwide, CXR has become prevalent in routine examinations and pre-employment screenings region-wide.

However, the analysis of chest radiographs requires years of training and experience. An X-ray projection produces a 2D image in which anatomical structures can obscure one another and abnormal features of various pathologies may be difficult to differentiate. A study in the United Kingdom found that the difference in reading accuracy between a general practitioner and a radiologist specialist can be significant, with concordance scores between $58\%$ and $74.6\%$ depending on years of medical practice \cite{satia2013assessing}. In Thailand, tens of millions of chest X-ray images are produced annually, yet there are only 1,277 radiologists in the entire country \cite{tmcstat2019}. It would be extremely difficult and cost-intensive to scale the number of qualified radiology specialists to meet the ever-growing demand for accurate CXR interpretation.

Recent advances in machine learning have introduced a wide variety of computer vision methods for medical image analysis, and deep learning has become a dominant trend in computer-aided diagnosis for chest X-rays. Several studies have demonstrated strong classification performance by convolutional neural networks \cite{rajpurkar2017chexnet, irvin2019chexpert, singh2018deep, guan2018diagnose}, supported by a wealth of open datasets released by medical centers around the world \cite{wang2017nihdataset, irvin2019chexpert, johnson2019mimic, bustos2019padchest}. However, several studies suggest that these accuracies may not transfer to unseen datasets, particularly when algorithms trained on one population are applied to another \cite{sathitratanacheewin2018deep}. Overfitting and limited generalizability remain major obstacles to real-world deployment.

In this work, we present the development and comprehensive validation of the chest radiograph analysis model in Inspectra CXR version 5. Building on the finding that local adaptation is central to accuracy on Thai populations \cite{chamveha2020local}, this model is designed around three properties that matter for real clinical use. First, it performs \emph{simultaneous classification and localization}: alongside multi-label classification, it produces anatomically focused heatmaps that indicate \emph{where} a suspected lesion is, using an architecture that augments a DenseNet-121 backbone \cite{huang2017densely} with Attend-and-Compare Modules (ACM) \cite{kim2020acm} and a Probabilistic Class Activation Map (PCAM) aggregation layer \cite{ye2020pcam}. Second, it is validated for \emph{generalization across sites}, using a dedicated independent test set drawn from 13 hospitals across Thailand in addition to a large in-domain test set. Third, we evaluate the system for \emph{clinical validation} through a usability study in which thoracic radiologists assess the concordance of the system's classification and localization with their own reading and rate the overall system using the System Usability Scale (SUS). This model was developed on 874,858 frontal chest radiographs from Siriraj Hospital, the largest such development effort we have undertaken.

Our contributions are threefold:
\begin{enumerate}
    \item We describe a large-scale, localization-capable CXR system trained on 874,858 frontal radiographs with radiologist-verified labels, and report its performance across nine clinically important conditions.
    \item We quantify both in-domain accuracy (mean AUROC $0.994$ on 19,871 held-out cases) and cross-site generalization (mean AUROC $0.970$ on 5,992 cases from 13 hospitals), along with lesion-level localization performance (mean LLF $77.9\%$).
    \item We report a clinical usability evaluation with five thoracic radiologists, demonstrating high classification and localization concordance (93.6\% and 94.7\%) and a SUS score of 89, indicating strong real-world acceptability.
\end{enumerate}

\section{Related Work}

\subsection{Deep Learning for Chest X-ray Classification}
The release of large public chest radiograph datasets such as ChestX-ray14 \cite{wang2017nihdataset}, CheXpert \cite{irvin2019chexpert}, MIMIC-CXR \cite{johnson2019mimic}, and PadChest \cite{bustos2019padchest} has driven rapid progress in automated abnormality detection. Early influential work such as CheXNet \cite{rajpurkar2017chexnet} demonstrated that deep convolutional networks could approach radiologist-level performance on individual findings such as pneumonia, and subsequent studies extended these results to multi-label settings across many thoracic conditions \cite{singh2018deep, guan2018diagnose, guendel2018learning}. Densely connected architectures \cite{huang2017densely} and residual networks \cite{he2016deep} are common backbones for these tasks, often pretrained on ImageNet \cite{deng2009imagenet}.

Despite these advances, generalization remains a central concern. Models trained on one institution or population frequently degrade when applied elsewhere, a phenomenon attributed to differences in imaging hardware, acquisition protocols, disease prevalence, and labeling conventions \cite{sathitratanacheewin2018deep}. Our prior study \cite{chamveha2020local} quantified this effect specifically for the Thai population and showed that local adaptation closes much of the gap. The present work continues this line by validating generalization explicitly across a network of Thai hospitals rather than at a single center.

\subsection{Weakly Supervised Localization}
Whole-image classification labels are far more abundant than pixel-level annotations, motivating weakly supervised approaches that localize findings using only image-level supervision. Class Activation Mapping (CAM) \cite{zhou2016learning} and its variants are widely used to produce saliency heatmaps, but standard CAM heatmaps are often diffuse and poorly aligned with lesion boundaries. The Probabilistic Class Activation Map (PCAM) pooling method \cite{ye2020pcam} reformulates the aggregation step to improve both the precision and the lesion coverage of the resulting heatmaps while preserving classification accuracy. Complementarily, the Attend-and-Compare Module (ACM) \cite{kim2020acm} explicitly contrasts features from different image regions --- mirroring how radiologists compare areas of a radiograph --- and has been shown to benefit medical image recognition. Our system combines both mechanisms so that the same model that classifies a study also yields clinically interpretable localization.

\subsection{Clinical Validation of CXR AI}
Beyond retrospective accuracy, a growing body of work emphasizes prospective and reader-centric evaluation. Multi-reader multi-case (MRMC) studies have shown that comprehensive CXR models can improve radiologist accuracy when used as assistance \cite{seah2021effect}, and AI-aided interpretation has been reported to improve both reader performance and reporting efficiency \cite{ev_C3_12_ref, ev_P3_7}. Standardized MRMC protocols for evaluating such assistance have been published \cite{ev_E1_19}, and dedicated detection systems for malignant pulmonary nodules and lung cancer have been developed and validated against radiologist reads, including in screening populations \cite{nam2019development, ev_P1_35, ev_E5_1}. Prospective deployments of AI triage systems in real clinical workflows have also begun to appear \cite{ev_E1_20}. In high-tuberculosis-burden settings --- directly relevant to Thailand and Southeast Asia --- deep learning detectors have matched radiologist performance and been evaluated for triage across multiple algorithms and sites \cite{ev_E1_36, ev_E4_17}. Several commercial CXR systems, including solutions from Lunit, Qure.ai, Carebot, and Chang Gung, have entered clinical investigation \cite{ev_C2_6, ev_C3_3, ev_C3_12, ev_C3_16, ev_C3_16_ref}, reflecting the maturation of the field toward regulated clinical use. A locally relevant evaluation of multi-finding CXR detection has also been reported from Thailand \cite{ev_E2_8}. Usability, in addition to accuracy, is increasingly recognized as a determinant of successful adoption; the System Usability Scale \cite{brooke2013sus, bangko1986sus} provides a validated, widely used instrument for quantifying perceived usability. In this work we assess both reading concordance and SUS-based usability with practicing thoracic radiologists, complementing the technical validation.

\section{Methodology}

\subsection{Data Acquisition}
Our model development dataset consists of chest X-ray images and corresponding radiologist reports collected from Siriraj Hospital, Bangkok, Thailand, between 2010 and 2019. The dataset contains a total of $874{,}858$ frontal chest radiographs with paired radiologist reports. All images were de-identified, and the use of this dataset was approved by the Siriraj Institutional Review Board (SIRB).

To ensure data quality and clinical relevance, we applied explicit inclusion and exclusion criteria. To be included, an image had to be a frontal posteroanterior (PA) view acquired with digital radiography, belong to a patient aged 18 years or older, provide adequate lung-field coverage, have a complete and accessible radiologist report, and demonstrate adequate penetration, positioning, and contrast for diagnostic interpretation. We excluded non-standard projections (lateral, anteroposterior portable, oblique or rotated, and lordotic or apical views), images with quality issues (motion artifacts, blurring, over- or under-exposure, or equipment malfunction), and images with data-integrity problems (missing or corrupted DICOM metadata, or duplicate examinations from the same patient on the same date).

\subsection{Data Quality and Annotation}
We implemented a multi-step data quality assurance protocol. Automated checks on DICOM metadata and reports first verified the digital-radiography requirement, confirmed that all subjects were $\geq 18$ years old, and validated the projection type so that only frontal PA views were retained. Proprietary image-processing algorithms then assessed positioning accuracy and overall diagnostic quality, flagging images with technical deficiencies. Comprehensive file-system checks confirmed the existence and completeness of the corresponding radiologist report for each retained image.

Labels were generated through a two-stage process. First, initial labels for radiographic abnormalities were extracted from the free-text radiologist reports using a proprietary natural-language labeling model. Board-certified chest radiologists then reviewed and corrected these potentially noisy labels to ensure accuracy and clinical relevance, producing multi-label classification targets for each image. Localization is learned in a weakly supervised manner from image-level labels alone; no pixel-level annotations are used in training. Pixel-level annotations were produced solely to evaluate localization: experienced chest radiologists annotated the location and boundaries of pathological findings on a held-out subset of abnormal cases, forming the localization test set (Dataset-C) described in Section \ref{sec:development}.

The system supports nine clinically important conditions: atelectasis, cardiomegaly, pulmonary edema, lung opacity, mass, nodule, pleural effusion, tuberculosis, and pneumothorax.

\subsection{Development and Test Datasets} \label{sec:development}
The final dataset comprised $874{,}858$ cases, all with image-level classification labels. A dedicated test set of $19{,}871$ images was held out and never used during training or model selection; the remaining images formed the development dataset, which was split into training and validation partitions. Every label in the held-out test sets was comprehensively reviewed and confirmed by board-certified chest radiologists, providing definitive ground-truth annotations against which performance was measured.
Table~\ref{table:dev_dist} shows the distribution across conditions for the entire data, consisting of the combined training, validation, and test sets.

\begin{table}[htbp]
\begin{center}
\caption{Dataset distribution per condition. Counts are multi-label (a study may carry more than one condition).}
\begin{tabular}{|c|c|c|c|}
\hline
\textbf{Condition} & \textbf{Number of Cases}\\
\hline
Lung Opacity & $170{,}007$\\
\hline
Mass & $31{,}590$\\
\hline
Nodule & $81{,}587$\\
\hline
Pulmonary Edema & $18{,}670$\\
\hline
Atelectasis & $37{,}268$\\
\hline
Cardiomegaly & $212{,}991$\\
\hline
Pleural Effusion & $51{,}675$\\
\hline
Tuberculosis &$29{,}108$\\
\hline
Pneumothorax & $472$\\
\hline
\end{tabular}
\label{table:dev_dist}
\end{center}
\end{table}

To evaluate the system, we defined three test datasets, summarized in Table~\ref{table:datasets}. Dataset-A is the in-domain classification test set held out from the Siriraj development data. Dataset-B is an independent \emph{generalization} set drawn from 13 hospitals across Thailand, used to assess cross-site robustness. Dataset-C is the localization test set with radiologist pixel-level annotations.

\begin{table}[htbp]
\begin{center}
\caption{Test datasets used for the algorithm Verification and Validation (V\&V) report.}
\begin{tabular}{|c|c|l|}
\hline
\textbf{Dataset} & \textbf{\# cases} & \textbf{Description}\\
\hline
Dataset-A & $19{,}871$ & In-domain (Siriraj) classification test set\\
\hline
Dataset-B & $5{,}992$ & Generalization classification set from 13 Thai hospitals\\
\hline
Dataset-C & $4{,}549$ & Localization test set with radiologist annotations\\
\hline
\end{tabular}
\label{table:datasets}
\end{center}
\end{table}

The per-condition distribution of the test sets is given in Table~\ref{table:class_dist}: Dataset-A and Dataset-B report classification cases, while Dataset-C reports localization annotations (applicable to six conditions for which heatmaps are evaluated).

\begin{table}[htbp]
\begin{center}
\caption{Distribution per condition across the test sets. Dataset-A (in-domain) and Dataset-B (generalization) report classification cases; Dataset-C reports localization annotations. ``--'' denotes conditions not evaluated for localization.}
\begin{tabular}{|c|c|c|c|}
\hline
\textbf{Condition} & \textbf{Dataset-A} & \textbf{Dataset-B} & \textbf{Dataset-C}\\
\hline
Lung Opacity & $4{,}061$ & $763$ & $1{,}137$\\
\hline
Mass & $975$ & $98$ & $1{,}710$\\
\hline
Nodule & $2{,}013$ & $321$ & $606$\\
\hline
Pulmonary Edema & $573$ & $142$ & --\\
\hline
Atelectasis & $1{,}164$ & $118$ & $623$\\
\hline
Cardiomegaly & $5{,}019$ & $767$ & --\\
\hline
Pleural Effusion & $1{,}181$ & $335$ & $450$\\
\hline
Tuberculosis & $1{,}933$ & $1{,}080$ & --\\
\hline
Pneumothorax & $64$ & $47$ & $177$\\
\hline
Normal & $11{,}242$ & $3{,}281$ & --\\
\hline
\end{tabular}
\label{table:class_dist}
\end{center}
\end{table}

\subsection{Model Architecture}
\begin{figure}[htbp]
\centering
\includegraphics[width=\linewidth]{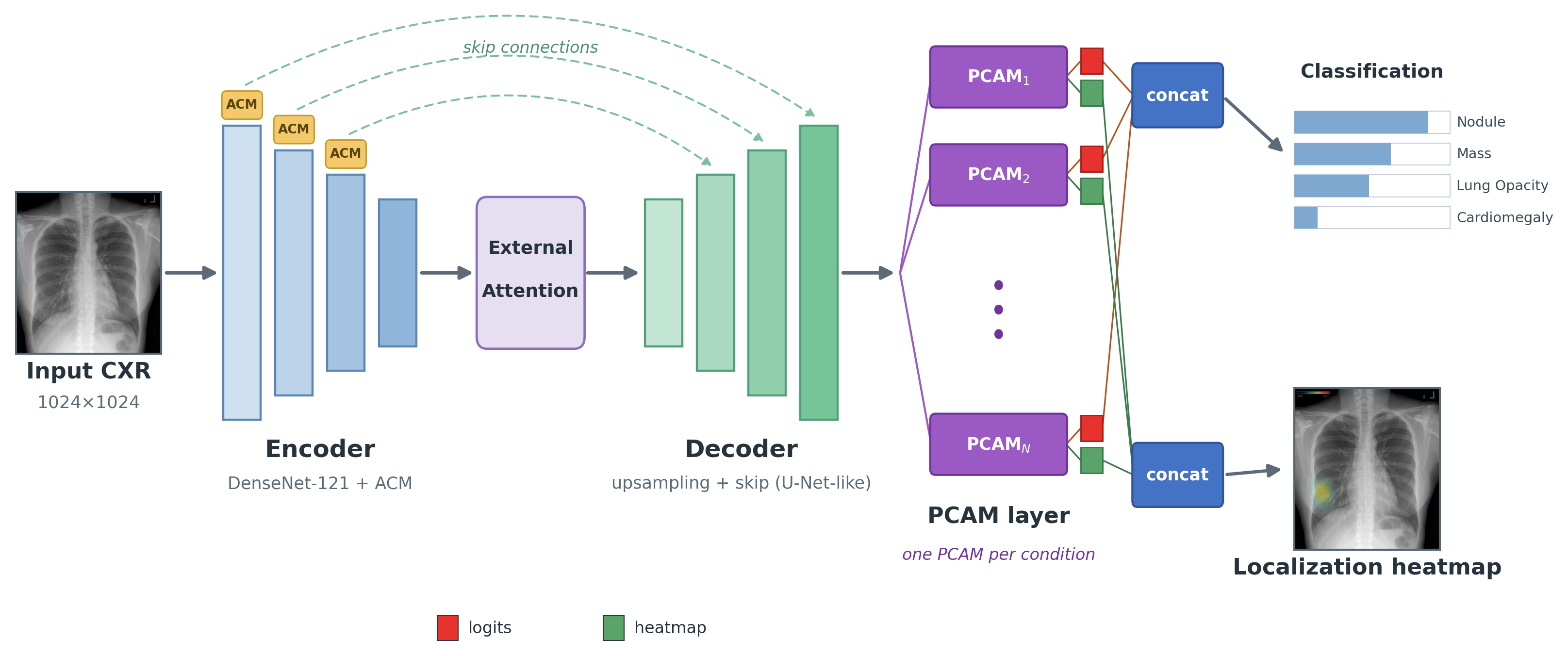}
\caption{Overall structure of the Inspectra CXR version 5 analysis model: a DenseNet-121 + ACM encoder, an External Attention stage, and a decoder with upsampling and skip connections (similar to a U-Net), followed by a PCAM (+ Cut-Noise) aggregation layer with one branch per condition. Each PCAM branch emits a class logit and a per-condition heatmap; the logits are concatenated into the classification output and the heatmaps into the localization output. The model is trained in a multi-task fashion for simultaneous image-level classification and pixel-level localization. The input and localization examples shown are from a de-identified case.}
\label{fig:architecture}
\end{figure}

The model is trained in a multi-task fashion to perform image-level multi-label classification and pixel-level lesion localization simultaneously. It comprises an encoder, a decoder, and an aggregation layer, so that spatial information is preserved throughout and lesion locations can be recovered (Figure~\ref{fig:architecture}).

\paragraph{Encoder (DenseNet-121 + ACM).} The encoder backbone is a DenseNet-121 network \cite{huang2017densely} pretrained on ImageNet \cite{deng2009imagenet}, augmented with Attend-and-Compare Modules (ACM) \cite{kim2020acm} inserted after convolutional blocks. Rather than a conventional attention mechanism, the ACM layer explicitly compares different regions of the input image with their surrounding context, mirroring how radiologists compare symmetric or semantically related lung zones to identify subtle abnormalities. Input images are resized to $1024 \times 1024$ pixels to preserve the fine detail needed to detect small lesions such as nodules or early tuberculosis.

\paragraph{Decoder with External Attention.} An External Attention layer learns correlations across the dataset, and the decoder then employs upsampling layers with skip connections --- similar to a U-Net architecture --- to combine high-level semantic features with low-level spatial detail, supporting precise localization of abnormal findings.

\paragraph{Aggregation (PCAM + Cut-Noise).} For feature aggregation we use a Probabilistic Class Activation Map (PCAM) pooling layer \cite{ye2020pcam}, a global pooling operation that leverages the localization ability of class activation mapping while producing heatmaps that are more precise and provide greater lesion coverage. This is enhanced with a Cut-Noise technique that improves the clarity and reliability of the generated heatmaps by attenuating spurious activations outside lesion regions. The aggregation layer emits, for each condition, both a classification score and a heatmap, yielding simultaneous classification and localization.

\subsection{Training Configuration}
Images were resized to $1024 \times 1024$ pixels and normalized with histogram equalization. The model was initialized with ImageNet weights and trained for 10 epochs with a batch size of 10, using the Adam optimizer with an initial learning rate of $0.0001$ reduced by a factor of $0.1$ at epochs 7 and 9. The training objective was binary cross-entropy with per-class weighting to address class imbalance.

\subsection{Operating Point Selection}
After training, we computed the positive predictive value, sensitivity, and specificity on the validation set. For each condition, we selected the operating point on the validation set to balance sensitivity and specificity at a clinically appropriate positive predictive value for that finding.

\subsection{Evaluation Metrics}
\paragraph{Classification.} For each condition we report accuracy, sensitivity, specificity, and the area under the receiver operating characteristic curve (AUROC), computed against radiologist-verified labels on Dataset-A and Dataset-B.

\paragraph{Localization.} On Dataset-C we evaluate the predicted heatmaps against radiologist annotations using two intersection-based overlap measures. The intersection over ground truth (IoGT) is the area of overlap between the predicted heatmap and the annotated lesion divided by the lesion area, reflecting how much of the true lesion the heatmap covers (a recall-like measure). The intersection over heatmap (IoHM) is the same overlap divided by the heatmap area, reflecting how much of the highlighted region actually falls on the lesion (a precision-like measure). We further report free-response operating characteristic statistics at an IoU threshold of $0.5$: the lesion localization fraction (LLF), the proportion of true lesions correctly localized, and the non-lesion localization fraction (NLF), the average number of spurious localizations per image.

\paragraph{Clinical usability.} We additionally assess the system with practicing radiologists through (i) a concordance analysis comparing the system's outputs with the radiologists' reading for both classification and localization, and (ii) the System Usability Scale \cite{brooke2013sus}.

The \emph{concordance test} measures how closely the system agrees with radiologists and is conducted separately for classification and for localization. In both cases each study is assigned to one of four categories --- Agree, Edit, Add, or Reject --- and the first three are counted as \emph{concordant} while Reject is counted as \emph{discordant}; the concordance rate is the proportion of concordant studies.

For \emph{classification concordance}, the findings are first reduced to a common set of categories: because several conditions are used interchangeably to describe pulmonary opacities, lung opacity, mass, nodule, pulmonary edema, and atelectasis are combined into a single ``Lung Opacity Group'', with cardiomegaly, pleural effusion, tuberculosis, and pneumothorax kept separate. Each study is then assigned to one of the categories in Table~\ref{table:concordance_cls}.

\begin{table}[htbp]
\begin{center}
\caption{Classification concordance categories.}
\begin{tabular}{|l|p{8.2cm}|c|}
\hline
\textbf{Category} & \textbf{Definition} & \textbf{Result}\\
\hline
Agree & The radiologist report matches the system completely & Concordant\\
\hline
Edit & The report is in partial agreement with the system & Concordant\\
\hline
Add & System flagged at least one finding, and the radiologist adds further findings & Concordant\\
\hline
Reject & System and radiologist disagree: a false alert on a normal study, a normal finding on an abnormal study, or an incorrect finding & Discordant\\
\hline
\end{tabular}
\label{table:concordance_cls}
\end{center}
\end{table}

For \emph{localization concordance}, the system's heatmaps are compared with the radiologist-annotated lesion locations (lesion type is not considered, as the heatmap is a single overlay for all findings), using the categories in Table~\ref{table:concordance_loc}.

\begin{table}[htbp]
\begin{center}
\caption{Localization concordance categories.}
\begin{tabular}{|l|p{8.2cm}|c|}
\hline
\textbf{Category} & \textbf{Definition} & \textbf{Result}\\
\hline
Agree & Ground-truth boxes overlap the heatmap by $>50\%$ for all lesions & Concordant\\
\hline
Edit & The heatmap covers all lesions but also highlights some non-lesion regions & Concordant\\
\hline
Add & The heatmap misses some lesions & Concordant\\
\hline
Reject & No overlap between the heatmap and the lesion areas & Discordant\\
\hline
\end{tabular}
\label{table:concordance_loc}
\end{center}
\end{table}

Finally, SUS scores are obtained from five thoracic radiologists who used the system for an extended period and rated ten standard statements on a five-point scale, with the final score computed following the standard SUS procedure.

\section{Experiments and Results}

\subsection{In-Domain Classification Performance}
On the held-out in-domain test set (Dataset-A, $19{,}871$ radiologist-verified cases), the model achieved strong performance across all nine conditions, with a mean AUROC of $0.994$, mean sensitivity of $92.35\%$, and mean specificity of $98.55\%$ (Table~\ref{table:results_a}). Performance was uniformly high, with per-condition AUROC ranging from $0.983$ (tuberculosis) to $1.000$ (pulmonary edema and pleural effusion).

\begin{table}[htbp]
\begin{center}
\caption{Classification performance on the in-domain test set (Dataset-A, $19{,}871$ cases).}
\begin{tabular}{|c|c|c|c|c|}
\hline
\textbf{Condition} & \textbf{Accuracy} & \textbf{Sensitivity} & \textbf{Specificity} & \textbf{AUROC}\\
\hline
Lung Opacity & $95.49\%$ & $90.15\%$ & $97.42\%$ & $0.991$\\
\hline
Mass & $98.89\%$ & $89.44\%$ & $99.71\%$ & $0.999$\\
\hline
Nodule & $96.48\%$ & $92.30\%$ & $97.23\%$ & $0.991$\\
\hline
Pulmonary Edema & $99.77\%$ & $95.81\%$ & $99.97\%$ & $1.000$\\
\hline
Atelectasis & $98.51\%$ & $91.07\%$ & $99.28\%$ & $0.998$\\
\hline
Cardiomegaly & $95.69\%$ & $89.48\%$ & $98.46\%$ & $0.994$\\
\hline
Pleural Effusion & $99.59\%$ & $97.63\%$ & $99.80\%$ & $1.000$\\
\hline
Tuberculosis & $94.76\%$ & $91.36\%$ & $95.34\%$ & $0.983$\\
\hline
Pneumothorax & $99.69\%$ & $93.94\%$ & $99.71\%$ & $0.986$\\
\hline
\textbf{Mean} & $\mathbf{97.65\%}$ & $\mathbf{92.35\%}$ & $\mathbf{98.55\%}$ & $\mathbf{0.994}$\\
\hline
\end{tabular}
\label{table:results_a}
\end{center}
\end{table}

\subsection{Cross-Site Generalization}

To assess robustness beyond the development institution, we evaluated the model on Dataset-B, an independent set of $5{,}992$ cases collected from 13 hospitals across Thailand. The model retained strong discrimination, with a mean AUROC of $0.970$ (Table~\ref{table:results_b}) --- only $0.024$ below the in-domain mean of $0.994$. Per-condition AUROC remained high across sites (e.g. $0.983$ for atelectasis), indicating that the model separates diseased from non-diseased cases nearly as well on external data as in-domain and that its ranking ability generalizes across heterogeneous imaging hardware, acquisition protocols, and case mix.

Mean sensitivity ($82.22\%$) and specificity ($95.89\%$) shifted more than AUROC, with the change concentrated in sensitivity; the largest reductions were for atelectasis ($66.94\%$) and pulmonary edema ($77.40\%$), both diffuse, low-contrast findings. Because AUROC is threshold-independent while sensitivity and specificity are evaluated at a single operating point, this is the expected behavior when a fixed, in-domain-selected threshold is applied to a population with different score distributions: the preserved AUROC shows the shift reflects where the threshold falls rather than degraded separability. The most-affected conditions also have few positives in the multi-site set (118 atelectasis, 142 pulmonary edema), making their per-condition sensitivity estimates imprecise. This behavior is consistent with cross-site sensitivity reductions reported for other commercial CXR software \cite{ev_E1_43, ev_E1_56}.

\begin{figure}[htbp]
\centering
\includegraphics[width=\linewidth]{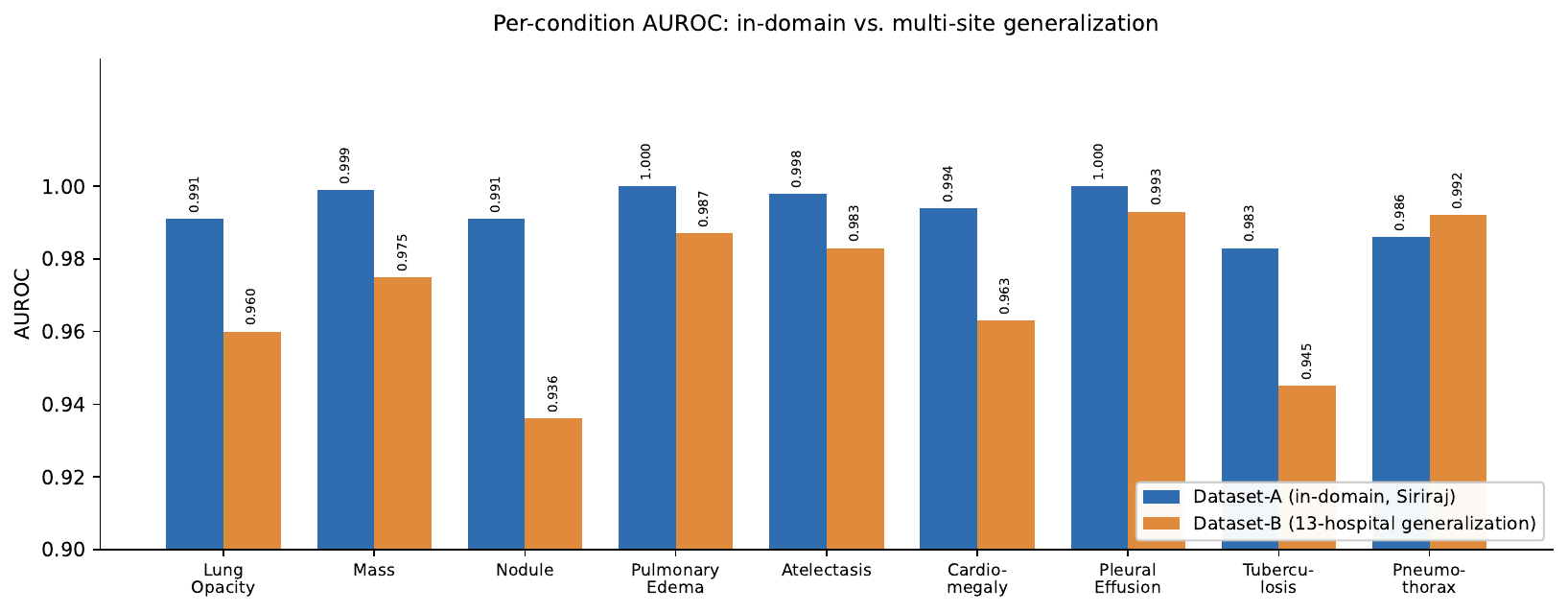}
\caption{Per-condition AUROC on the in-domain test set (Dataset-A) and the multi-site generalization set (Dataset-B). Performance remains high across sites, with the largest AUROC gaps for nodule and tuberculosis.}
\label{fig:auroc}
\end{figure}

Figure~\ref{fig:auroc} compares per-condition AUROC between the in-domain and multi-site test sets.

\begin{table}[htbp]
\begin{center}
\caption{Classification performance on the multi-site generalization set (Dataset-B, $5{,}992$ cases from 13 hospitals).}
\begin{tabular}{|c|c|c|c|c|}
\hline
\textbf{Condition} & \textbf{Accuracy} & \textbf{Sensitivity} & \textbf{Specificity} & \textbf{AUROC}\\
\hline
Lung Opacity & $91.81\%$ & $87.23\%$ & $93.51\%$ & $0.960$\\
\hline
Mass & $94.46\%$ & $82.52\%$ & $95.00\%$ & $0.975$\\
\hline
Nodule & $92.07\%$ & $74.70\%$ & $94.56\%$ & $0.936$\\
\hline
Pulmonary Edema & $98.23\%$ & $77.40\%$ & $99.56\%$ & $0.987$\\
\hline
Atelectasis & $97.21\%$ & $66.94\%$ & $98.86\%$ & $0.983$\\
\hline
Cardiomegaly & $90.57\%$ & $86.24\%$ & $92.02\%$ & $0.963$\\
\hline
Pleural Effusion & $97.63\%$ & $90.72\%$ & $98.64\%$ & $0.993$\\
\hline
Tuberculosis & $90.54\%$ & $84.91\%$ & $92.55\%$ & $0.945$\\
\hline
Pneumothorax & $98.17\%$ & $89.36\%$ & $98.30\%$ & $0.992$\\
\hline
\textbf{Mean} & $\mathbf{94.52\%}$ & $\mathbf{82.22\%}$ & $\mathbf{95.89\%}$ & $\mathbf{0.970}$\\
\hline
\end{tabular}
\label{table:results_b}
\end{center}
\end{table}

\subsection{Lesion Localization Performance}
We evaluated the quality of the model's heatmaps on Dataset-C, comprising $4{,}549$ radiologist-annotated cases across six conditions for which heatmap localization is applicable. Because the model receives no pixel-level supervision during training, this localization performance is achieved entirely through weak supervision from image-level labels. As shown in Table~\ref{table:results_c}, the model achieved a mean lesion localization fraction (LLF) of $77.90\%$ at an IoU threshold of $0.5$, with only $0.58$ non-lesion localizations per image (NLF) on average, indicating that the heatmaps are both sensitive to true lesions and reasonably specific. Localization was strongest for lung opacity (LLF $89.21\%$) and weakest for pneumothorax (LLF $63.64\%$), the latter being a thin, peripheral finding that is intrinsically difficult to delineate. The low IoHM for nodule ($0.13$) is expected rather than a localization failure: because IoHM divides the overlap by the heatmap area, a fixed-scale heatmap necessarily over-covers a small lesion and drives the precision-like ratio down, while the recall-like LLF for nodule remains high at $83.41\%$.

Beyond indicating whether an abnormality is present, these heatmaps show \emph{where} it lies, which can accelerate radiologist verification and support trust in the system's output \cite{ev_P5_30}.

\begin{table}[htbp]
\begin{center}
\caption{Localization performance on Dataset-C ($4{,}549$ annotated cases) at IoU $0.5$. LLF: lesion localization fraction; NLF: non-lesion localizations per image.}
\begin{tabular}{|c|c|c|c|c|}
\hline
\textbf{Condition} & \textbf{IoGT} & \textbf{IoHM} & \textbf{LLF} & \textbf{NLF}\\
\hline
Lung Opacity & $0.4400$ & $0.6171$ & $89.21\%$ & $1.23$\\
\hline
Mass & $0.6264$ & $0.5702$ & $84.41\%$ & $0.21$\\
\hline
Nodule & $0.8466$ & $0.1316$ & $83.41\%$ & $0.94$\\
\hline
Atelectasis & $0.5460$ & $0.3660$ & $66.43\%$ & $0.40$\\
\hline
Pleural Effusion & $0.6661$ & $0.3958$ & $80.33\%$ & $0.16$\\
\hline
Pneumothorax & $0.4195$ & $0.3249$ & $63.64\%$ & $0.59$\\
\hline
\textbf{Mean} & $\mathbf{0.5908}$ & $\mathbf{0.4009}$ & $\mathbf{77.90\%}$ & $\mathbf{0.58}$\\
\hline
\end{tabular}
\label{table:results_c}
\end{center}
\end{table}

\subsection{Clinical Usability Evaluation}
Beyond retrospective accuracy, we evaluated how the system performs in the hands of practicing radiologists. Five certified thoracic radiologists participated, and the system was used by the participants over more than two months before responses were collected, giving participants substantial hands-on experience.

\paragraph{Concordance.} We measured the agreement between the system's outputs and radiologists for both classification and localization (Table~\ref{table:concordance}). For classification, evaluated on the $19{,}871$ cases of Dataset-A, $18{,}596$ cases were concordant (Agree, Edit, or Add) and $1{,}275$ were discordant (Reject), yielding a classification concordance rate of $93.60\%$. For localization, evaluated on $4{,}549$ cases, $4{,}309$ were concordant and $240$ discordant, yielding a localization concordance rate of $94.72\%$. The Add category warrants comment, since it corresponds to the system under-calling relative to the radiologist. We count it as concordant because it reflects the radiologist supplementing, rather than contradicting, the system's findings. The clinically consequential error is Reject: for classification, the system and radiologist indicate conflicting conditions; for localization, the heatmap shows no overlap with any true lesion. Reject is therefore the only category counted as discordant.

\begin{table}[htbp]
\begin{center}
\caption{Reading concordance between Inspectra CXR version 5 and radiologists.}
\begin{tabular}{|c|c|c|c|c|c|c|}
\hline
\multirow{2}{*}{\textbf{Task}} & \multicolumn{4}{c|}{\textbf{\# cases}} & \multirow{2}{*}{\textbf{Total}} & \multirow{2}{*}{\textbf{Concordance}}\\
\cline{2-5}
& \textbf{Agree} & \textbf{Edit} & \textbf{Add} & \textbf{Reject} & &\\
\hline
Classification & $15{,}737$ & $2{,}003$ & $856$ & $1{,}275$ & $19{,}871$ & $93.60\%$\\
\hline
Localization & $789$ & $2{,}767$ & $753$ & $240$ & $4{,}549$ & $94.72\%$\\
\hline
\end{tabular}
\label{table:concordance}
\end{center}
\end{table}

\paragraph{System Usability Scale.} The five radiologists rated the system using the standard ten-item SUS questionnaire. The mean SUS score was $89$, well above the acceptance threshold of $68$ and corresponding to an ``excellent'' rating on standard SUS interpretation scales \cite{bangko1986sus}. This indicates that the system is not only accurate but also perceived as highly usable in routine practice. User feedback gathered during the evaluation consisted entirely of usability improvement suggestions, with no functional defects reported.

\section{Discussion}

We have shown that a locally developed, localization-capable chest radiograph model can combine high in-domain accuracy (mean AUROC $0.994$) with robust cross-site generalization (mean AUROC $0.970$ across 13 hospitals), reliable lesion localization (mean LLF $77.9\%$), and strong clinical acceptance (classification and localization concordance of $93.6\%$ and $94.7\%$, and an ``excellent'' SUS score of $89$). Together these results indicate that the system is accurate, generalizes across heterogeneous Thai hospitals, and is trusted by practicing radiologists as an assistive second reader. The main remaining limitations concern subtle or small findings --- particularly small nodules and lower-zone or plate atelectasis --- and occasional over-diffuse heatmaps, which are the focus of ongoing model and data improvements. Per-condition ROC analysis from case-level prediction scores is planned as future work.

\section*{Conclusion}

We have presented the development and comprehensive validation of the chest radiograph analysis model in Inspectra CXR version 5, a deep learning model for thoracic disease detection on chest radiographs in Thailand. Trained on $874{,}858$ frontal radiographs from Siriraj Hospital, the system achieved a mean AUROC of $0.994$ in-domain and $0.970$ across 13 hospitals, a mean lesion localization fraction of $77.9\%$, classification and localization concordance of $93.6\%$ and $94.7\%$ with radiologists, and a System Usability Scale score of $89$.

These results indicate that a locally developed, localization-capable CXR system can simultaneously deliver high accuracy, generalize across heterogeneous Thai hospitals, and earn the trust of practicing radiologists. These results support the use of a locally developed model as a reliable algorithmic second reader that highlights likely-abnormal cases and their locations for radiologist confirmation.

The road toward safe AI adoption in medicine requires proper validation with sufficiently large and relevant datasets, guidance from medical professionals, and attention to how end users interact with the system.
Future work will focus on the failure modes identified in this evaluation, namely subtle findings, lesions obscured by bony structures, and over-diffuse heatmaps, and on further improving generalizability across the diversity of Thai hospitals through additional training data.
\bibliography{references,references_evidence}{}
\bibliographystyle{unsrt}

\end{document}